\pdfoutput=1

\documentclass[11pt]{article}

\usepackage{authblk}
\usepackage[]{emnlp2021}

\usepackage{times}
\usepackage{latexsym}

\usepackage[T1]{fontenc}

\usepackage[utf8]{inputenc}

\usepackage{microtype}

\usepackage{multirow}
\usepackage{graphicx}
\usepackage[euler]{textgreek}
\usepackage{textcomp}
\usepackage{mwe}
\usepackage{amsmath}
\DeclareMathOperator*{\argmax}{argmax}

\definecolor{darkgreen}{rgb}{0.0, 0.4, 0.13}

\title{"It doesn't look good for a date": Transforming Critiques into Preferences for Conversational Recommendation Systems}

\author[1]{{\bf Victor S. Bursztyn}}
\author[2]{{\bf Jennifer Healey}}
\author[2]{{\bf Nedim Lipka}}
\author[2]{\\{\bf Eunyee Koh}}
\author[1,3]{{\bf Doug Downey}}
\author[1]{{\bf Larry Birnbaum}\vspace{-1.5ex}}
\affil[1]{Department of Computer Science, Northwestern University, Evanston, IL, USA}
\affil[2]{Adobe Research, San Jose, CA, USA}
\affil[3]{Allen Institute for Artificial Intelligence, Seattle, WA, USA}
\affil[ ]{\texttt{v-bursztyn@u.northwestern.edu,\,\{jehealey,lipka,eunyee\}@adobe.com}}
\affil[ ]{\texttt{\{d-downey,l-birnbaum\}@northwestern.edu}}

\begin{document}
\maketitle


\begin{abstract}
Conversations aimed at determining good recommendations are iterative in nature. People often express their preferences in terms of a critique of the current recommendation (e.g., ``It doesn't look good for a date''), requiring some degree of common sense for a preference to be inferred. In this work, we present a method for transforming a user critique into a positive preference (e.g., ``I prefer more romantic'') in order to retrieve reviews pertaining to potentially better recommendations (e.g., ``Perfect for a romantic dinner''). We leverage a large neural language model (LM) in a few-shot setting to perform critique-to-preference transformation, and we test two methods for retrieving recommendations: one that matches embeddings, and another that fine-tunes an LM for the task. We instantiate this approach in the restaurant domain and evaluate it using a new dataset of restaurant critiques. In an ablation study, we show that utilizing critique-to-preference transformation improves recommendations, and that there are at least three general cases that explain this improved performance.
\end{abstract}

\section{Introduction}

Conversational recommendation systems (CRSs) are dialog-based systems that aim to refine a set of options over multiple turns of a conversation, envisioning more natural interactions and better user modeling than in non-conversational approaches.

However, the resulting dialogs still do not necessarily reflect how real conversations unfold. Most CRSs fall into two categories: they either frame the problem as a slot-filling task within a predefined feature space, such as \citet{sun2018conversational, zhang2018towards, budzianowski-etal-2018-multiwoz}, which is closer to how people make decisions but not as flexible as real conversations; or they elicit preferences by asking users to rate specific items, such as \citet{christakopoulou2016towards}, which is independent of a feature space but not as natural to users.

\begin{figure}[htb]
  \centering
  \includegraphics[width=.5\textwidth]{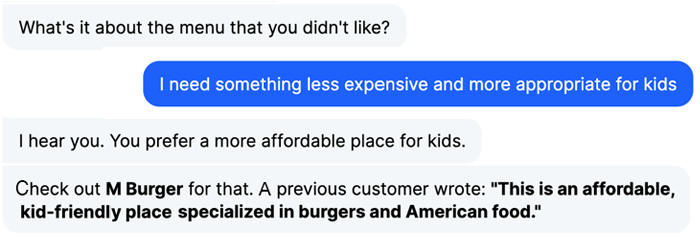}
  \caption{An example of our system transforming a critique into a positive preference and then using a customer testimonial to sell the user on a new option.}
  \label{fig:example}
\end{figure}

When we examine situations involving real human agents \cite{lyu2021workflow}, decisions typically require multiple rounds of recommendations by the agent and critiques by the user, with the agent continuously improving the recommendations based upon user preferences that can be inferred from such critiques.

These inferences can be compared to the types of common sense inferences that have been studied recently with LMs \cite{davison-etal-2019-commonsense, majumder-etal-2020-like, jiang2021m}. However, use of LMs for critique interpretation remains underexplored, despite the important role of critiques in communicating preferences---a very natural real-world task. Working in the restaurant domain, we prompt GPT3 \cite{brown2020language} to transform a free-form critique (e.g., ``It doesn't look good for a date'') into a positive preference (e.g., ``I prefer more romantic'') that better captures the user's needs. Compared with most previous work on common sense inference, which relies on manually-constructed question sets, our task presents an opportunity to study common sense inference within a naturally arising, real-world application.

We test the effect of our novel critique interpretation method on the quality of recommendations using two different methods: one that matches the embedding of an input statement (e.g., ``I prefer more romantic'') to persuasive arguments found in customer reviews (e.g., ``Perfect for a romantic dinner''); and another one that fine-tunes BERT \cite{devlin2018bert} in using an input statement to rank a given set of arguments.

Our work differs from previous critiquing-based systems that strongly limit the types of critiques that can be used \cite{chen2012critiquing} and aligns with a recent trend in the CRS literature towards more open-ended interactions \cite{radlinski2019coached, byrne2019taskmaster}. To the best of our knowledge, \citet{penha2020does} are the closest prior work investigating whether BERT can be used for recommendations by trying to infer related items and genres. Here, we focus specifically on critique-to-preference inferences, aiming at more natural dialogs and better recommendations.

Our contributions are the following: 1. We propose a critique interpretation method that does not limit the feature space a priori; 2. We demonstrate that transforming critiques into preferences improves recommendations over two fold when matching embeddings and by 19-59\% when fine-tuning an LM to rank recommendations, and present three possible explanations for this; and 3. We release a new dataset of user critiques in the restaurant domain, contributing a new applied task where common sense has great practical value.

\section{Methods}

In this section, we describe three methods: A critique interpretation method (\ref{critique-understanding}), an embeddings-based recommender (\ref{recommendation-search}), and an LM-based recommender (\ref{recommendation-ranking}).

\subsection{Critique Interpretation}
\label{critique-understanding}

Critique interpretation is the task of transforming a free-form critique into a positive preference. Our critique interpretation method uses GPT3 in a few-shot setting similarly to \citet{brown2020language}, which can be represented in a 3-shot version as follows:

\small
\label{eq:1}
\begin{align*}
&{\rm \bf GPT3\,Input:} \\
&It\,looks\,cheap=>I\,prefer\,a\,fancier\,place.\, \\
&Too\,expensive=>I\,prefer\,a\,more\,affordable\,place.\, \\
&That's\,so\,tacky=>I\,prefer\,a\,more\,stylish\,place. \\
&That's\,not\,good\,for\,a\,date\;=>\;I\,prefer \\
&\\
&{\rm \bf GPT3\,Output:} \\
&a\,more\,romantic\,place.
\nonumber
\end{align*}
\normalsize

To prime GPT3 for our task, we include ten examples in its prompt, five related to food and five to the atmosphere.\footnote{Fully available at \url{https://bit.ly/3fnf8V2}} We then append the critique that we would like to transform followed by the string \textit{``I prefer''}, which conditions GPT3 to generate a positive preference. In our experiments, positive preferences are sampled using OpenAI's Completion API (the DaVinci model, temperature = 0.7, top p = 1.0, response length = 20, and no penalties).

Besides not requiring a hand-crafted feature set, this method is also capable of more flexible interpretation of language, such as transforming ``How come they only serve that much?''---with no clearly negative words---into ``I prefer larger portions.''

\subsection{Content-based Recommendations}

\subsubsection{Recommendation Search}
\label{recommendation-search}

Our embeddings-based recommender, $f_{cos}$, takes a preference statement and searches for persuasive arguments in customer reviews. As seen in Figure \ref{fig:example}, we can define a persuasive argument as a review sentence that conveys clearly positive sentiment while being as specific as possible w.r.t the user's preferences.

To incorporate this definition in $f_{cos}$, first we parse sentences in customer reviews using spaCy \cite{honnibal2017spacy} and use EmoNet \cite{abdul2017emonet} to keep the sentences with at least a minimum amount of ``joy'' ($\geq 0.7$) as our set of argument candidates $A$.

Then we use the Universal Sentence Encoder \cite{cer2018universal} to calculate the similarity of all these argument candidates w.r.t a given user preference. We calculate the cosine similarity between their representations in this embedding space, select the argument with maximum alignment, and recommend the associated restaurant:

\small
\begin{gather*}
  Sim(s_1, s_2) = cos(Enc(s_1), Enc(s_2))\\
  f_{cos}(preference) = \argmax_{a \in A} Sim(a, preference)
\nonumber
\end{gather*}
\normalsize

As with critique interpretation, $f_{cos}$ can take any natural language statement as input to search for potential recommendations. We denote $f_{cos}^{pref}$ when it uses an inferred positive preference as input (``I prefer more romantic'') and $f_{cos}^{crit}$ when it directly uses a critique (``It doesn't look good for a date''). In our first ablation study, we use $f_{cos}^{crit}$ as a baseline to test the efficacy of $f_{cos}^{pref}$ in retrieving better recommendations. 

\subsubsection{Recommendation Ranking}
\label{recommendation-ranking}

Besides using pretrained embeddings to search for recommendations from customer reviews, we design a more computationally intensive method, $f_{LM}$, that fine-tunes BERT to rank a set of arguments $A$ considering a given input statement.

We use the currently top performing open-source solution \cite{han2020learning, TensorflowRankingKDD2019} on the MSMARCO passage ranking leaderboard\footnote{\url{https://microsoft.github.io/msmarco/}} to fine-tune three versions of BERT: $f_{LM}^{pref}$ uses a positive preference as input (``I prefer more romantic''), $f_{LM}^{crit}$ uses a critique (``It doesn't look good for a date''), and $f_{LM}^{concat}$ uses a concatenation of both a critique and a preference (``It doesn't look good for a date. I prefer more romantic''). Hypothetically, the more powerful LM method could learn to satisfy the user's preferences without the need of critique interpretation if the performances of $f_{LM}^{crit} \approx f_{LM}^{pref} \approx f_{LM}^{concat}$.

In our experiments, BERT-Base is fined-tuned for 10,000 steps, with learning rate = $10^{-5}$, maximum sequence length = 512, and softmax loss, using a Nvidia Quadro RTX 8000 for 3-6h per run (when ranking 15 and 30 arguments, respectively) and two runs per model (2-fold cross validation).

\section{Evaluation}

We run two ablation studies to evaluate the hypothesis that critique interpretation would be beneficial to the overall recommendation approach. First, we analyze our embeddings-based recommender, $f_{cos}$, to check whether the performance of $f_{cos}^{pref} > f_{cos}^{crit}$. Secondly, we analyze our LM fine tuning-based recommender, $f_{LM}$, to check if $f_{LM}^{pref} > f_{LM}^{crit}$ or $f_{LM}^{concat} > f_{LM}^{crit}$. Finally, we discuss qualitative differences between the tested arms.

\begin{table*}[ht]
\scalebox{0.67}{
\small
\centering
\begin{tabular}{|l|l|l|l|l|l|l|l|}
\hline
\multicolumn{1}{|c|}{\multirow{2}{*}{Test case}}                     & \multicolumn{1}{c|}{\multirow{2}{*}{Positive preference}}                 & \multicolumn{3}{c|}{Without critique interpretation ($f_{cos}^{crit}$)}                                                                                                                                                                                                                                            & \multicolumn{3}{c|}{With critique interpretation($f_{cos}^{pref}$)}                                                                                                                                                                                                                                                                       \\ \cline{3-8} 
\multicolumn{1}{|c|}{}                                               & \multicolumn{1}{c|}{}                                                     & Rank \#1                                                                                                             & Rank \#2                                                                                           & Rank \#3                                                                                        & Rank \#1                                                                                                                         & Rank \#2                                                                                                  & Rank \#3                                                                                                        \\ \hline
\begin{tabular}[c]{@{}l@{}}It looks\\ too casual.\end{tabular}       & \begin{tabular}[c]{@{}l@{}}I prefer a\\ fancier place.\end{tabular}   & \begin{tabular}[c]{@{}l@{}}Very cheesy,\\ very fresh!\end{tabular}                                                   & Very kid friendly.                                                                                 & \begin{tabular}[c]{@{}l@{}}Awesome\\ ambiance!\end{tabular}                                     & \textbf{\begin{tabular}[c]{@{}l@{}}Elegant, upscale\\ and classy place\\ for a special occasion.\end{tabular}}                   & \textbf{\begin{tabular}[c]{@{}l@{}}The best\\ restaurant\\ around here.\end{tabular}} & \textbf{\begin{tabular}[c]{@{}l@{}}Superior restaurant,\\ the only place I\\ will have a dim sum.\end{tabular}}              \\ \hline
\begin{tabular}[c]{@{}l@{}}It has a\\ freaking band!\end{tabular}    & \begin{tabular}[c]{@{}l@{}}I prefer a more\\ quiet place.\end{tabular} & \begin{tabular}[c]{@{}l@{}}It has an awesome\\ atmosphere.\end{tabular}                                              & \begin{tabular}[c]{@{}l@{}}It has an awesome\\ atmosphere.\end{tabular}                            & \begin{tabular}[c]{@{}l@{}}It has a great\\ atmosphere.\end{tabular}                            & \textbf{\begin{tabular}[c]{@{}l@{}}Excellent spot to\\ spend time alone\\ or talk business.\end{tabular}}                                                                 & \begin{tabular}[c]{@{}l@{}}Good ambiance.\end{tabular} & \begin{tabular}[c]{@{}l@{}}Great place\\ to be at night.\end{tabular} \\ \hline
\begin{tabular}[c]{@{}l@{}}I don't really\\ like seafood.\end{tabular} & \begin{tabular}[c]{@{}l@{}}I prefer beef\\ or chicken.\end{tabular}    & \begin{tabular}[c]{@{}l@{}}Everything delicious\\ with an exception of\\ of the shrimps.\end{tabular} & \begin{tabular}[c]{@{}l@{}}I found that I do not\\enjoy tuna, but my mom\\thought it was excellent.\end{tabular} & \begin{tabular}[c]{@{}l@{}}For dinner, I enjoyed\\ the scallops one night\\ and the sea bass\\the second.\end{tabular} & \textbf{\begin{tabular}[c]{@{}l@{}}I only eat Beef\\Brisket here because\\is delicious!\end{tabular}} & \textbf{\begin{tabular}[c]{@{}l@{}}Chicken flautas\\are always\\delish.\end{tabular}}                                & \textbf{\begin{tabular}[c]{@{}l@{}}Chicken moist\\and tender.\end{tabular}}            \\ \hline
\end{tabular}}
\caption{Three test cases with the top 3 arguments from $f_{cos}^{crit}$ and $f_{cos}^{pref}$ (accurate marked in bold).}
\label{tab:qualitative-critique-understanding}
\end{table*}
\normalsize

\subsection{Data}

Our methods were instantiated in a system comprising 15 restaurants selected from two of the largest metropolitan areas in the United States, covering a variety of price ranges and cuisines. For each restaurant, up to 100 four- or five-star customer reviews were collected from Google Places. This resulted in a total of 1455 reviews comprising 5744 sentences, 2865 of which pass the threshold for being identified as positive review sentences.

We compiled a set of user critiques from two sources: a set of 46 unique critiques from user studies that were conducted to test an earlier system prototype \cite{bursztyn2021}, and 294 additional critiques adapted from the Circa dataset \cite{louis2020d}. Circa was designed to study indirect answers to yes-no questions, such as ``Are you a big meat eater?'' answered with ''I prefer leafy greens'', from which the critique ``I'm not a big meat eater'' can be generated. We end with a total of 340 individual critiques after examining 1205 similar examples.

We generated a positive preference for each individual critique using our critique interpretation method in \ref{critique-understanding}, without discarding any critiques. Our method yielded accurate preferences for 298 critiques (87.6\%). For the remaining 42, we found GPT3 mostly undecided and vague (e.g., ``Jalapeños are my limit'' generates ``I prefer food without jalapeños''). In our experiments, for these edge cases, we kept the best of three trials, but we believe that results using just the first generation would have been qualitatively similar.

The 340 critiques were randomly combined into pairs and triples in order to simulate longer conversations, i.e., two- and three-round critiques. We sampled 340 pairs and 340 triples, substituting only exceptional combinations that contained contradictory statements (e.g., ``I'm not a big meat eater.'' paired with ``I'm not in the mood for vegetables.''), for a total of 1020 critiques. Compound critiques were concatenated into single statements as well as their corresponding preferences.

This curated dataset of 1020 restaurant critiques and inferred preferences is made available to the research community.\footnote{\url{https://github.com/vbursztyn/critique-to-preference-emnlp2021}}

\subsection{Measurements}
\label{offline}

For evaluating our embedding-based methods $f_{cos}$, we use critiques as input to $f_{cos}^{crit}$ and their positive preferences as input to $f_{cos}^{pref}$. For each query we retrieve the top 3 arguments, which are labeled as accurate or inaccurate by a human judge (illustrated in Table \ref{tab:qualitative-critique-understanding}). To measure labeling consistency, a second human annotator redundantly labeled a sample of 100 arguments resulting in a Cohen's Kappa of 0.71, which indicates strong agreement.

We then measure Precision@1, Precision@2, and Precision@3 in Table \ref{tab:exp1-results} for the embeddings-based method with ($f_{cos}^{pref}$) and without critique interpretation ($f_{cos}^{crit}$).

To train and evaluate the BERT-based method $f_{LM}$, we retrieve the top 15 arguments from $f_{cos}^{crit}$ and the top 15 arguments from $f_{cos}^{pref}$ for 100 queries. Each argument receives a score from 3 (very relevant) to 1 (irrelevant). Again, a second human annotator relabeled 100 arguments for a Cohen's Kappa of 0.73, also indicating strong agreement.

We design three ranking tasks: $task_1$ consists of ranking the 15 arguments originally retrieved with $f_{cos}^{crit}$, hence closer to critiques in the embedding space; $task_2$ consists of ranking the 15 arguments originally retrieved with $f_{cos}^{pref}$, hence closer to preferences; and $task_3$ consists of ranking both sets, i.e., 30 arguments. For each task we train $f_{LM}^{pref}$, $f_{LM}^{crit}$, and $f_{LM}^{concat}$. We then measure nDCG@1, nDCG@3, nDCG@5, and nDCG@10 in Table \ref{tab:exp2-results} averaged after 2-fold cross validation.

\begin{table}[]
\scalebox{0.80}{
\small
\begin{tabular}{|l|l|l|l|}
\hline
                             & Precision@1     & Precision@2     & Precision@3     \\ \hline
$f_{cos}^{crit}$ & 0.256          & 0.251          & 0.250          \\ \hline
$f_{cos}^{pref}$ & \textbf{0.574} & \textbf{0.546} & \textbf{0.525} \\ \hline
\end{tabular}}
\caption{Precision@1, 2, and 3 for $f_{cos}^{crit}$ and $f_{cos}^{pref}$.}
\label{tab:exp1-results}
\end{table}
\normalsize

\begin{table}[]
\scalebox{0.80}{
\small
\begin{tabular}{|l|l|l|l|l|l|}
\hline
                         & model              & nDCG1           & nDCG3           & nDCG5           & nDCG10          \\ \hline
\multirow{3}{*}{$task_1$} & $f_{LM}^{crit}$            & 0.617          & 0.674          & 0.723          & 0.811          \\ \cline{2-6} 
                         & \textbf{$f_{LM}^{pref}$}   & \textbf{0.731} & \textbf{0.753} & \textbf{0.773} & \textbf{0.858} \\ \cline{2-6} 
                         & $f_{LM}^{concat}$          & 0.726          & 0.740          & 0.773          & 0.844          \\ \hline
\multirow{3}{*}{$task_2$} & $f_{LM}^{crit}$            & 0.676          & 0.754          & 0.805          & \textbf{0.865} \\ \cline{2-6} 
                         & $f_{LM}^{pref}$            & 0.729          & 0.761          & 0.774          & 0.856          \\ \cline{2-6} 
                         & \textbf{$f_{LM}^{concat}$} & \textbf{0.805} & \textbf{0.772} & \textbf{0.808} & 0.863          \\ \hline
\multirow{3}{*}{$task_3$} & $f_{LM}^{crit}$            & 0.498          & 0.537          & 0.605          & 0.660          \\ \cline{2-6} 
                         & \textbf{$f_{LM}^{pref}$}   & \textbf{0.790} & \textbf{0.754} & \textbf{0.758} & \textbf{0.791} \\ \cline{2-6} 
                         & $f_{LM}^{concat}$          & 0.686          & 0.663          & 0.685          & 0.746          \\ \hline
\end{tabular}}
\caption{nDCG1, 3, 5, and 10 for $f_{LM}$ on each task.}
\label{tab:exp2-results}
\end{table}
\normalsize

\subsection{Results}

We found that using the positive preferences yields substantial improvements in information retrieval. For $f_{cos}$, in Table \ref{tab:exp1-results}, $f_{cos}^{pref}$ increases Precision@1 by \textbf{124\%}, Precision@2 by \textbf{118\%}, and Precision@3 by \textbf{110\%}. This gap is also present, with marginal variations, when separately analyzing single-, two-, and three-round critiques. For $f_{LM}$, in Table \ref{tab:exp2-results}, $f_{LM}^{pref}$ outperforms $f_{LM}^{crit}$ by \textbf{19\%} on nDCG@1 even at $task_1$, where $f_{LM}^{crit}$ could have an edge. This gap persists for $task_2$ ($f_{LM}^{concat}$ outperforms by \textbf{19\%}), increases for $task_3$ ($f_{LM}^{pref}$ outperforms by \textbf{59\%}), and tends to narrow towards nDCG@10. Overall, we found strong evidence in support of our hypothesis.

Table \ref{tab:qualitative-critique-understanding} shows three examples in which the use of positive preferences was clearly beneficial. These examples represent three critique patterns that cause systematic errors if critique interpretation is turned off: 1. When the user implies a preference for a feature using the polar opposite (e.g., ``It looks too casual'' implying ``I prefer a fancier place''); 2. When the user draws on common sense to express a preference (``It has a freaking band!'' implying ``I prefer a more quiet place''); and 3. When the user implies a filter within a set of related features (e.g., ``I don't really like seafood'' implying preference for alternatives in the meat category).

Analyzing the results of $f_{cos}^{pref}$ and $f_{cos}^{crit}$ for the 340 single-round critiques, we found 170 cases where $f_{cos}^{pref}$ outperformed $f_{cos}^{crit}$. Within these, 40 belong to the first pattern (24\%), 78 to the second (46\%), and 38 to the third (22\%).\footnote{Fully available at \url{https://bit.ly/33SCKva}} A common trait behind the three patterns is that critiques can be lexically very distinct from their corresponding preference statements, and critique interpretation helps to bridge this gap.

\section{Conclusion \& Future Work}

In this paper, we presented an open-ended approach to content-based recommendations for CRS. We developed a novel critique interpretation method that uses GPT3 to infer positive preferences from free-form critiques. We also developed two methods for retrieving recommendations: one that matches embeddings and another that fine-tunes BERT for the task. We ran two ablation studies to test if transforming critiques into positive preferences would yield better recommendations, confirming that it improves performance across both methods. Finally, we described three critique patterns that cause systematic errors in recommendation search if critique interpretation is turned off.

For future work, we will strive to use critiques to identify and remove unsuitable restaurants; we speculate that the sparsity of customer reviews generally makes it harder to ``rule out'' than to ``rule in.'' We will also study other issues such as when to ask clarification questions to resolve ambiguity in the scope of a critique.

\section*{Acknowledgements}

We would like to thank reviewers for their helpful feedback. This work was supported in part by gift funding from Adobe Research and by NSF grant IIS-2006851.

\bibliography{emnlp}
\bibliographystyle{acl_natbib}

\end{document}